\documentclass[conference]{IEEEtran}
\IEEEoverridecommandlockouts

\usepackage{cite}
\usepackage{amsmath,amssymb,amsfonts}
\usepackage{algorithmic}
\usepackage{graphicx}
\usepackage{caption}
\usepackage[table]{xcolor}
\usepackage{subcaption}

\usepackage{textcomp}
\usepackage{xcolor}
\usepackage{newtxtext,newtxmath}
\usepackage{booktabs}
\usepackage{siunitx}
\usepackage{gensymb}
\usepackage{bm}
\usepackage{subcaption}
\usepackage{url}         
\usepackage{hyperref}    
\usepackage{cleveref}
\crefname{figure}{Fig.}{Figs.}   
\crefname{table}{Table}{Tables}  
\usepackage{textcomp}
\def\BibTeX{{\rm B\kern-.05em{\sc i\kern-.025em b}\kern-.08em
    T\kern-.1667em\lower.7ex\hbox{E}\kern-.125emX}}

\newcommand{\DOD}{Trail-based Off-road Multimodal Dataset}
\newcommand{\DODshort}{TOMD}
\newcommand{\mycomment}[1]{}

\makeatletter
\providecommand\sf@counterlist{}
\makeatother
\begin{document}
\IEEEaftertitletext{\vspace{-1.5em}}
\title{\DODshort: A \DOD\ for Traversable Pathway Segmentation under Challenging Illumination Conditions}

\author{
    \IEEEauthorblockN{
        Yixin Sun\IEEEauthorrefmark{1},
        Li Li\IEEEauthorrefmark{2},
        Wenke E\IEEEauthorrefmark{1},
        Amir Atapour-Abarghouei\IEEEauthorrefmark{1},
        Toby P. Breckon\IEEEauthorrefmark{1}
    }
    \IEEEauthorblockA{\IEEEauthorrefmark{1}%
        Department of Computer Science, Durham University, UK\\
        \{yixin.sun, wenke.e, amir.atapour-abarghouei, toby.breckon\}@durham.ac.uk
    }
    \IEEEauthorblockA{\IEEEauthorrefmark{2}%
        Department of Engineering, King’s College London, UK\\
        li.8.li@kcl.ac.uk
    }
}

\maketitle
\begin{abstract}
Detecting traversable pathways in unstructured outdoor environments remains a significant challenge for autonomous robots, especially in critical applications such as wide-area search and rescue, as well as in incident management scenarios such as forest fires. Current datasets and models primarily focus on either urban environments or wide vehicle-traversable off-road tracks, leaving a substantial gap in tackling the complexities of trail-based off-road scenarios. To address this issue, we introduce the \DOD\ (\DODshort), a comprehensive dataset explicitly designed for narrow and unstructured trail-like environments. Our dataset features high-fidelity multimodal sensor data — including 128-channel LiDAR, stereo imagery, GNSS, IMU, and illumination measurements — collected through repeated runs across diverse environmental conditions. In addition, we propose a novel dynamic multiscale data fusion model for precise traversable pathway prediction in trail-like areas. The study investigates the impact of various fusion processes — early, cross, and mixed — on model performance under different illumination levels: low-light, normal ambient lighting, and bright conditions. The results highlight the effectiveness of our approach, variation in performance across illumination levels, and the potential applicability of the dataset in diverse environmental conditions.

Our work provides a valuable resource for advancing trail-based off-road navigation, and we openly publish our \DODshort\ at \url{https://github.com/yyyxs1125/TMOD} to establish a future benchmark in this research domain.

\end{abstract}


\section{Introduction}
\noindent
Detecting traversable zones in unstructured outdoor environments pose significant challenges for autonomous robots, particularly in critical real-world applications such as wide-area search and rescue, as well as incident management scenarios such as forest fires. Off-road scenarios, particularly on narrow trails, pose unique challenges, such as restricted pathway widths for vehicles, vegetation that obscures the boundaries of traversable areas, and highly variable illumination due to canopy shading. Addressing these challenges necessitates high-quality, reliable data from a mobile acquisition platform. Moreover, data-driven deep learning approaches further intensify these requirements, as they demand extensive and diverse off-road datasets to ensure robust and generalisable performance under varying real-world operational conditions.

\begin{figure}[t]
  \centering
  \begin{subfigure}[t]{1.0\linewidth} 
    \vspace{0pt} 
    \includegraphics[width=\linewidth]{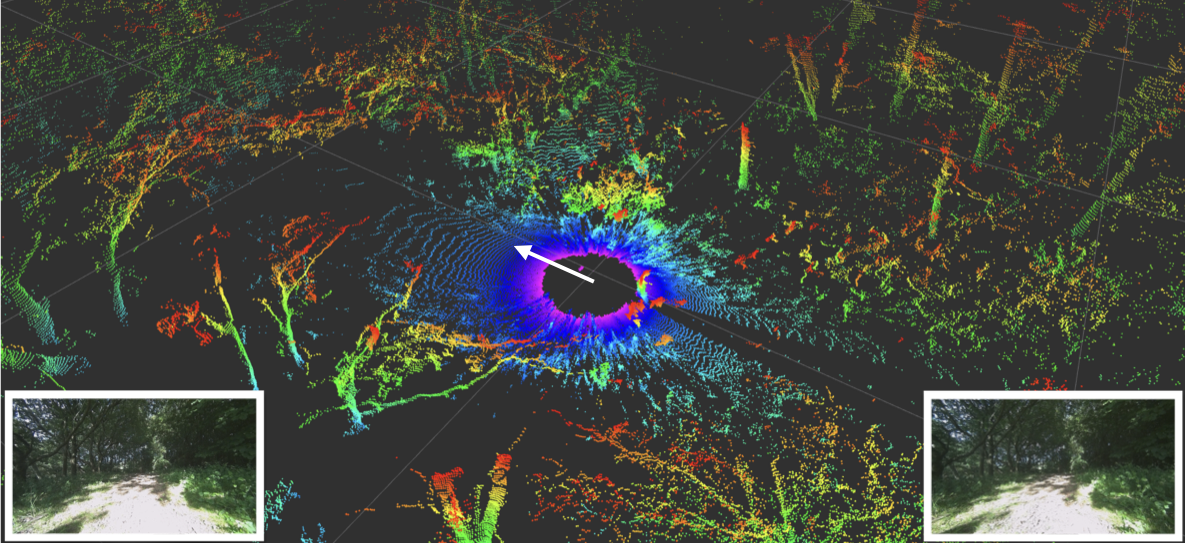}
    \caption{128-Channel LiDAR Point Clouds with corresponding left-lens (bottom-left) and right-lens (bottom-right) images from ZEDx Stereo camera. And the white arrow indicates the orientation of our robot.}
     \hfill
    \label{fig:short-a}
  \end{subfigure}
  
  \centering
  \begin{subfigure}[t]{0.48\linewidth} 
    \vspace{0pt} 
    \includegraphics[width=1.0\linewidth]{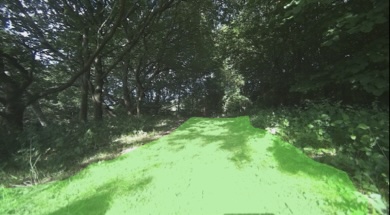}
    \caption{Image with annotated traver-sable trail ground truth (green). }
    \label{fig:}
  \end{subfigure}
  \begin{subfigure}[t]{0.48\linewidth} 
    \vspace{0pt} 
    \includegraphics[width=1.0\linewidth]{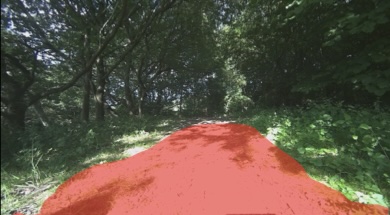}
    \caption{Image with predicted traver-sable trail (red).}
    \label{fig:}
  \end{subfigure}
  \caption{\DODshort~ overview: LiDAR and stereo camera visualization with ground truth and predicted traversable trail under high illumination.}
  \label{fig:overiew}
\end{figure}

Whilst existing datasets focused on urban environments\cite{geiger2013vision, sun2020scalability, caesar2020nuscenes} are widely used as benchmarks for road detection segmentation models, the development of datasets and models for off-road scenarios has lagged behind, with a significantly smaller catalogue of publicly available datasets, which predominantly focus on dual-track, vehicle-traversable off-road paths — referred to as `green lanes' in the UK and `jeep trails' in the US. These datasets largely overlook the narrower concept of human-traversable trails (or pathways), which, in reality, constitute the majority of off-road egress points into unstructured natural environments globally. To address this research gap, we introduce the \DOD, the first dataset explicitly designed to enhance perception and interpretation capabilities in trail environments based on the use of a medium scale all-terrain robot platform capable of transiting narrower off-road trails (An overview of \DODshort~is shown in Fig. \ref{fig:overiew}). 

The key contributions of our work in enhancing perception in such unstructured off-road environments are as follows:
\begin{itemize}
    \item The \DODshort, is specifically designed to represent complex unstructured trail-like scenarios using a medium-scale all-terrain robot platform that includes high-fidelity 3D LiDAR (128 channels), stereo imagery, GNSS, IMU, telemetry control data, and illumination measurements that are all collected via repeated route traversal under varying environmental conditions. In total, it comprises 31.4k frame pairs for image and LiDAR, along with key frame traversability level annotation.
    \item A novel data fusion-based dynamic multiscale model architecture is introduced for precise traversable pathway segmentation within such trail-like environments. The influence of different fusion processes — i.e., early, cross, and mixed fusion — on model performance is thoroughly examined under varying ambient illumination levels, including low, medium and high. This analysis demonstrates the potential applicability of \DODshort~in diverse environmental conditions.
    \item Calibrated and synchronized multi-sensor sequence data files, along with corresponding traversable-level RGB annotations and data processing tools, are publicly released as open-access resources to provide a new performance benchmark within the field of trail-based robotic navigation and autonomous exploration.
\end{itemize}
\section{Related Work}
\noindent We review prior work in two closely related areas: off-road datasets (Section~\ref{sec:existing}) and traversable area detection (Section~\ref{sec:TAD}).
\vspace{-0.2em}
\subsection{Existing Off-road datasets}
\label{sec:existing}
\begin{table*}[!ht]
\centering
\caption{Comparison of Existing Off-road Datasets featuring Sensor Modalities:- C: Camera, D: Depth Camera, G: GNSS, I: INS, L: LiDAR, M: IMU, N: NIR, U: Lux Meter. Camera resolution: width × height; LiDAR resolution: vertical channels.}
\footnotesize 
\setlength{\tabcolsep}{3pt} 
\rowcolors{2}{gray!10}{white}
\begin{tabular}{@{}>{\centering\arraybackslash}m{2.5cm} >{\centering\arraybackslash}m{1.2cm} >{\centering\arraybackslash}m{1.2cm} >{\centering\arraybackslash}m{1.2cm} >{\centering\arraybackslash}m{3cm} >{\centering\arraybackslash}m{2.5cm} >{\centering\arraybackslash}m{3cm} >{\centering\arraybackslash}m{2cm}@{}}
    \toprule
    Name & Sensors & Platform & Repeated & Anno-level (\#) & Camera Resolution & LiDAR Resolution & \#Frames (C/L) \\
    \midrule
    SOOR\cite{SOOR} & C & vehicle & No & objects (7) & 768$\times$384 & -- & 0.3k / - \\
    Wildscenes\cite{wildscenes} & C/G/I/L & handheld & No & objects (15) & 2016$\times$1512 & 16 & 9.3k / 12.1k \\
    YCOR\cite{YCOR} & C & vehicle & No & objects (8) & 1024$\times$544 & -- & 13.6k / - \\
    ORFD\cite{min2022orfd} & C/L & vehicle & No & traversability (2) & 1280$\times$720 & 40 & 12.2k / 12.2k \\
    CaT\cite{CaT} & C/I & vehicle & \textbf{Yes} & traversability (3) & 1024$\times$644 & -- & 12.3k / - \\
    GOOSE\cite{goose} & C/I/L & vehicle & \textbf{Yes} & objects (64) & 1024$\times$500 & \textbf{128} & 10.0k / 10.0k \\
    TartanDrive 2.0\cite{tartandrive} & C/G/I/L & vehicle & \textbf{Yes} & - & 1024$\times$512 & 32/70 &  250k / 250k\\
    Freiburgforest\cite{FreiburgForest} & C/D/N & robot & No & objects (6) & 1024$\times$768 & -- & 15.0k / - \\
    RUGD\cite{RUGD} & C & robot & No & objects (24) & 1920$\times$1200 & -- & 7.5k / - \\
    RELLIS-3D\cite{RELLIS-3D} & C/L/I & robot & No & objects (20) & 800$\times$592 & 64 & 6.2k / 13.6k \\
    Botanicgarden\cite{botanicgarden} & C/L/I & robot & No & objects (27) & 1920$\times$1200 & 16 & \textbf{2.3M / 1.2M} \\
    Finnwoodlands\cite{finnwoodlands} & C/L & handheld & No & objects (3) & 1280$\times$720 & 64 & 5.2k / 5.2k \\
    \textbf{\DODshort} \textbf{(Ours)} & C/G/L/M/\textbf{U} & robot & \textbf{Yes} & traversability (2) & 1920$\times$1080 & \textbf{128} & 31.4k / 31.4k \\
    \bottomrule
\end{tabular}
\label{tab:datasets}
\end{table*}
\noindent In comparison with urban autonomous driving datasets \cite{geiger2013vision, sun2020scalability, caesar2020nuscenes,li2021durlar}, off-road datasets have developed more slowly and exhibit significant gaps in terms of quantity and capacity.

\textbf{Collection platforms} influence the suitability of data collection methods, with robot-based and handheld-based approaches being more effective for capturing narrower trails/pathways (e.g., in forests or gardens) compared to vehicle-based methods. These trails are often narrower, more complex, unstructured, and characterised by dense vegetation. In some areas, dense tree canopies result in very low illumination conditions, posing challenges for camera-based data collection. Among these approaches, robot platform based data collection is preferred, as it both directly replicates realistic navigation/motion patterns and minimises any human-induced biases, leading to more consistent and representative data.

\textbf{Sensor modality} plays a critical role in datasets dedicated to unstructured, off-road driving scenarios. The first such dataset, RUGD \cite{RUGD}, utilised a Husky' platform robot equipped solely with a mono camera (Prosilica GT2750C) to capture video sequences, covering four general unstructured scenes: creek, park, trail, and village. However, the visual information provided by a single camera is insufficient for accurate path planning or prediction, as demonstrated by human-centric experiments in \cite{bauer2001relevance}. To address this limitation, an increasing number of datasets are expanding their sensor modalities to enrich the information available. For instance, CaT\cite{CaT} incorporates additional cameras to increase the field of view (FOV), and SOOR\cite{SOOR} integrates Freiburgforest\cite{FreiburgForest}, which introduces a depth camera and near-infrared (NIR) sensor. Additionally, \cite{RELLIS-3D, CaT, goose, botanicgarden, tartandrive} integrate Global Positioning System (GPS) and Inertial Measurement Unit (IMU) sensors to capture positional features. Most notably, \cite{YCOR, RELLIS-3D, min2022orfd, goose, wildscenes, tartandrive} adopt LiDAR, widely used in on-road datasets, to acquire spatial data that support environmental understanding.
\begin{figure*}[t]
  \centering
  \begin{subfigure}[t]{1.0\linewidth} 
    \vspace{0pt} 
    \includegraphics[width=\linewidth]{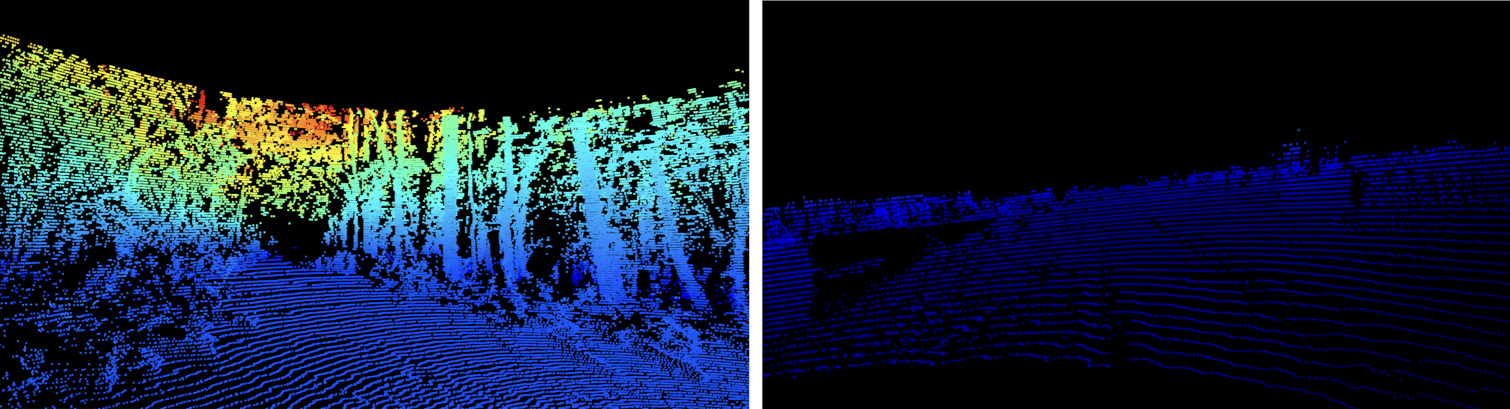}
    \caption{Exemplar LiDAR point clouds (left: \DODshort~(ours), 128 channels; right: RELLIS-3D, 64 channels.)}
     \hfill
    \label{fig:short-a}
  \end{subfigure}
  
  \centering
  \begin{subfigure}[t]{1.0\linewidth} 
    \vspace{0pt} 
    \includegraphics[width=1.0\linewidth]{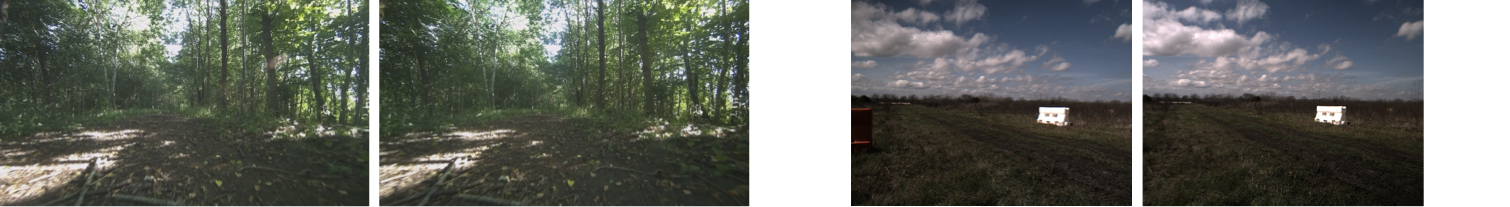}
    \caption{Stereo camera images (left pair: \DODshort~(ours), \num{1920}$\times$\num{1080}; right pair: RELLIS-3D, \num{800}$\times$\num{592}).}
    \label{fig:short-b}
  \end{subfigure}
  \caption{Visual Comparison Between \DODshort~ and the RELLIS-3D Dataset~\cite{RELLIS-3D}: (a) Point cloud from \DODshort~ with higher density, richer spatial information, and extended sensing range. (b) Wider stereo camera field of view in \DODshort~ compared to RELLIS-3D.}
  \label{fig:vis}
\end{figure*}

\textbf{High-vertical-resolution LiDAR} remains underutilised in off-road datasets, with no robot-based trail collections incorporating such sensors. Among handheld datasets, Wildscenes \cite{wildscenes} and Finnwoodlands \cite{finnwoodlands} use 16- and 64-channel LiDAR, respectively. Botanicgarden \cite{botanicgarden} and RELLIS-3D \cite{RELLIS-3D}, collected by robots, also employ 16- and 64-channel LiDAR. In vehicle-based collections, ORFD \cite{min2022orfd} uses a 40-channel LiDAR, TartanDrive 2.0 \cite{tartandrive} combines 32- and 70-channel LiDAR for wider viewpoints, and GOOSE \cite{goose} is the first to adopt 128-channel LiDAR, significantly enhancing spatial detail. Solid-State and Digital LiDAR (e.g., \cite{finnwoodlands, RELLIS-3D}) are preferred over Mechanical Spinning types (e.g., \cite{botanicgarden, goose, min2022orfd, wildscenes}) due to reduced point cloud distortion via simultaneous channel capture. The Ouster 128 LiDAR used in our dataset acquires all channels concurrently, ensuring high precision and consistency. A feature summary is provided in Table~\ref{tab:datasets}, with a visual comparison to RELLIS-3D shown in \cref{fig:vis}.

\begin{figure}[t]
  \centering
    \includegraphics[width=1.0\linewidth]{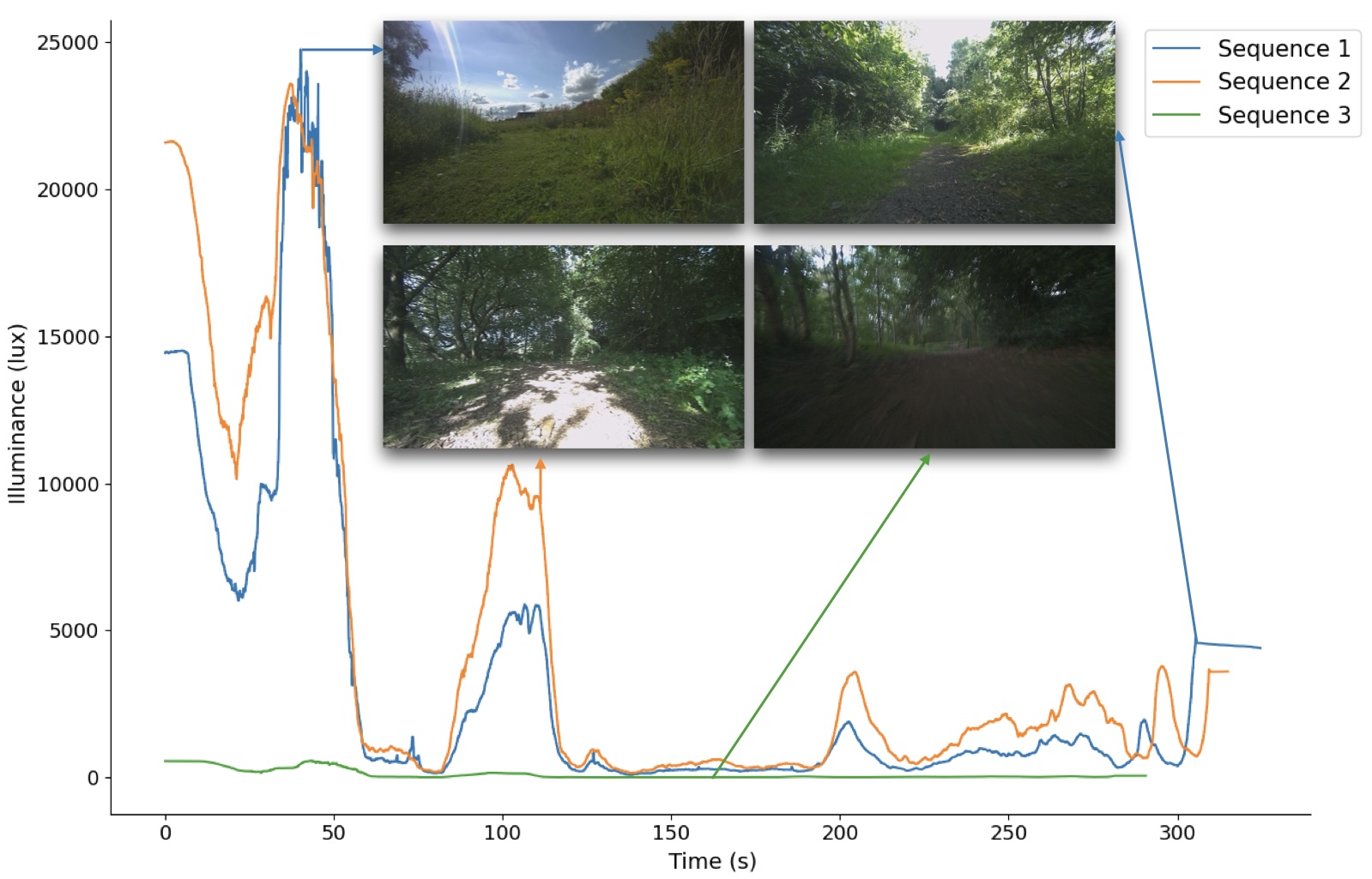}
   \caption{Illuminance trends over time for three exemplar sequences (sequence 1: afternoon, sequence 2: midday, sequence 3: dusk) from \DODshort~the corresponding left image camera views from the ZEDx stereo camera (inset). Excessively low illumination conditions can cause the camera to become underexposed, resulting in high levels of sensor noise and the loss of critical visual scene information.}
   \label{fig:lux}
\end{figure}

\textbf{Illumination intensity} contributes to dataset diversity by capturing data across different times, weather conditions, and seasons. However, most datasets \cite{SOOR, wildscenes, YCOR, min2022orfd, FreiburgForest, RUGD, RELLIS-3D, botanicgarden, finnwoodlands} do not include repeated paths under varying lighting. To address this, we integrate a lux meter to quantify ambient illumination (in lux) along identical routes. \cref{fig:lux} shows illuminance trends for three \DODshort\ sequences, alongside representative camera views for context.

\vspace{-0.2em}
\vspace{-0.2em}
\subsection{Traversable Area Detection}
\label{sec:TAD}
\noindent
Detecting the traversable area is typically formulated as a semantic segmentation problem, where the task is to predict regions navigable by autonomous vehicles. Recent advancements\cite{liang2022bevfusion,xiao2018hybrid} in on-road navigation suggest that fusing cameras and LiDAR data could be a potential solution to improve performance by mitigating the limitations of individual sensors. Based on the fusion stage, methods can be categorised into early fusion\cite{holder2018encoding,chen2019progressive}, late fusion\cite{gu2021cascaded}, and cross fusion \cite{caltagirone2019lidar}. Early fusion combines multiple input modalities at the beginning of processing, before feature extraction, which would allow the model to learn from all data sources simultaneously. In contrast, late fusion processes each modality separately through independent feature extraction pipelines and integrates them at a later stage, typically during the decision or prediction phase. Cross fusion, on the other hand, enables the interaction and exchange of information between different modalities at multiple stages throughout the processing pipeline, facilitating better integration and enhancing overall performance. While most methods have been developed and validated using the large-scale on-road KITTI dataset \cite{geiger2013vision}, the few that attempt to apply transfer learning from urban road scenes \cite{holder2016road} struggle to capture the complexities and variabilities of unstructured environments. Consequently, research specifically addressing off-road scenarios remains limited. OFF-Net\cite{min2022orfd} proposes a cross-attention-based model that dynamically fuses RGB data with surface normals derived from sparse LiDAR points. However, the assumption that traversable zones share similar surface normals may not always hold, as vegetation and other obstacles can cause irregularities, making the surface deviate from a typical on-road or vehicle traversable plane. 
\vspace{-0.2em}
\section{\DOD}
\label{sec:formatting}
\noindent
Compared to existing off-road datasets, \DODshort\ offers the following novel features:
\begin{itemize}
\item A high-vertical-resolution LiDAR with 128 channels, which eliminates the rolling shutter effect.
\item The first off-road dataset to utilise a lux meter for recording ambient illuminance.
\item Coverage of repeated trail-based routes under diverse environmental conditions.
\item The inclusion of recorded teleoperation commands provides detailed control-level information, which is valuable for route and traversability planning.
\item Integration of high-precision real-time kinematic (RTK) GNSS data to enable centimeter-level accuracy for localisation and mapping.
\end{itemize}
\subsection{Equipment and Sensor Setup}
\noindent
The \DOD\ is collected using the Rover Pro 4WD Robot, an all-terrain robot platform designed to withstand diverse environments and weather conditions. The Rover Pro has dimensions of 62.0 cm~$\times$~39.0 cm~$\times$~25.4 cm and a maximum speed of 2.5 m/s which is perfectly suited for activities in areas where conventional large-wheelbase (road) vehicles cannot readily access the terrain. 

\begin{figure}[t]
  \centering
    \includegraphics[width=1.0\linewidth]{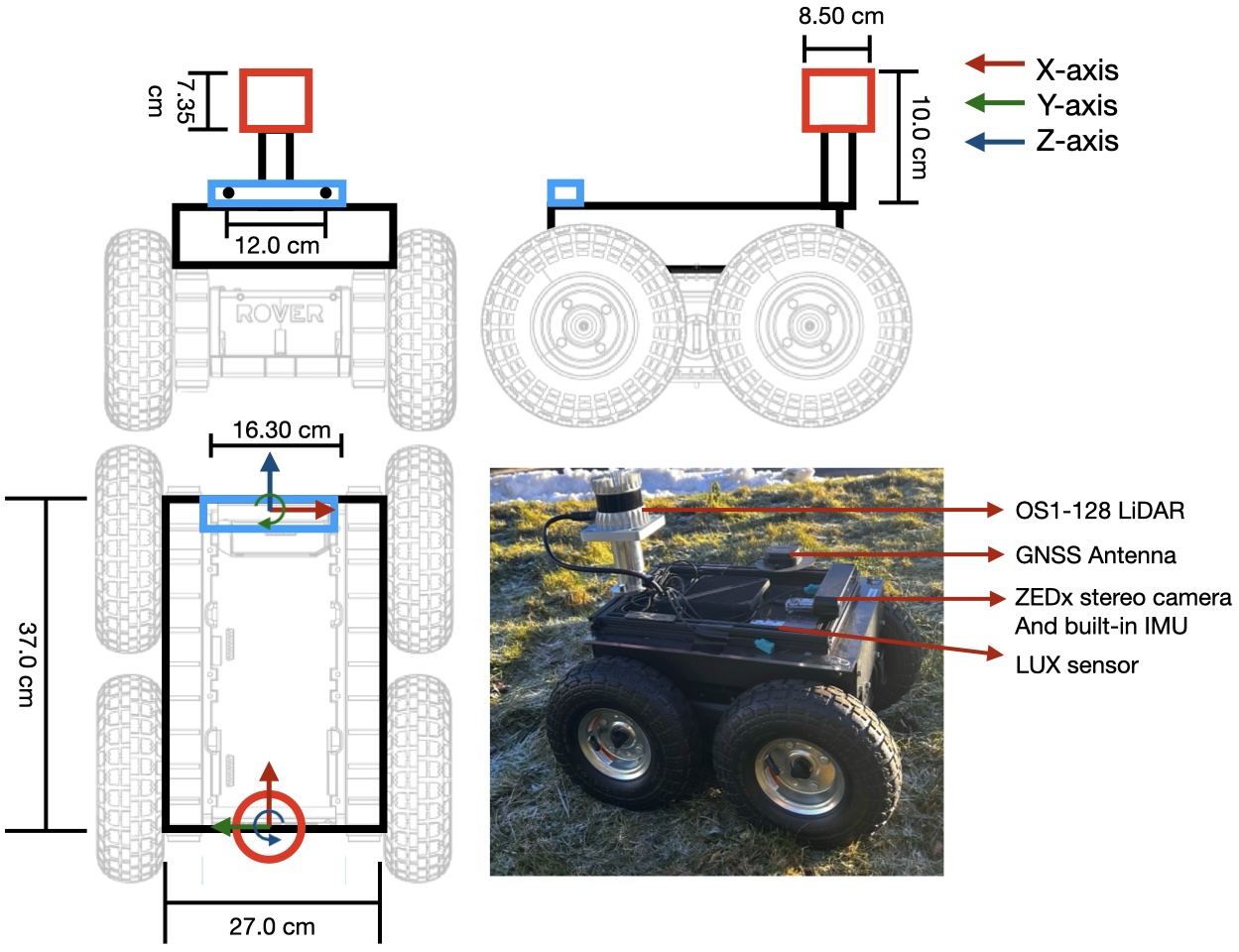}
    \caption{The Rover Pro robot is illustrated - front (top-left), top (bottom-left), and side (top-right) views, alongside a real-world image (bottom-right) indicating sensor mounting positions. Coordinate axes adhere to the right-hand rule.}
   \label{fig:robot}
\end{figure}

The robot is equipped with the following sensors, mounted on a custom water-resistant payload (see \cref{fig:robot}) to ensure reliability in challenging outdoor scenarios:
\begin{itemize}
    \item \textbf{LiDAR:} Ouster OS1 (128 channels) with a 865~\si{nm} laser wavelength, offering detection ranges of 100~\si{m} at \textgreater 90\% probability and 120~\si{m} at \textgreater50\% probability (under 100~\si{klx} sunlight, 80\% Lambertian reflectivity, 2048 points @ 10~\si{Hz}). Features include 0.3~\si{cm} range resolution, 360\si{\degree} horizontal FoV, and 45\si{\degree} vertical FoV.
    
    \item \textbf{Stereo Camera:} ZEDx dual-lens stereo camera with secure GMSL2 connection, designed for robust robotics use. Supports resolutions of 2~$\times$~(1920~$\times$~1200) @ 60~\si{fps} and 2~$\times$~(960~$\times$~600) @ 120~\si{fps}, with a maximum FoV of 110\si{\degree}~(H)~$\times$~80\si{\degree}~(V)~$\times$~120\si{\degree}~(D).
    
    \item \textbf{IMU:} Integrated IMU in the ZEDx camera comprising a 16-bit triaxial accelerometer and gyroscope. Provides $\pm$12~\si{G} accelerometer range with 0.36~\si{mg} resolution, $\pm$1000~\si{dps} gyroscope range with 0.03~\si{dps} resolution, and $\pm$0.5\% sensitivity error, at 400~\si{Hz} output rate.
    
    \item \textbf{GNSS:} ZED-F9P-0xB module with multi-band GNSS and RTK, embedded in the ZEDx NVIDIA Jetson Orin NX onboard computer. Offers up to 20~\si{Hz} update rate and 0.01~\si{m}~$\pm$~1~ppm positional accuracy (CEP).
    
    \item \textbf{Lux Meter:} Yoctopuce Light V4 USB ambient light sensor with 0.01~\si{lux} resolution, capable of measuring up to 83,000~\si{lux} at 10~\si{Hz}.
\end{itemize}

\noindent
A portable mini-PC powered by a NVIDIA Jetson Orin NX 16GB moduleserves as the onboard computer, chosen for its low power consumption. All sensors function as slaves and communicate with the onboard PC, acting as the master, via a standard ROS-based architecture. Technically, the onboard PC runs the ROS core to subscribe to topics containing data published by sensors, which operate as ROS nodes with precise time-stamping, and subsequent synchronisation, from a common software clock (ROS-based).

\subsection{Data Description}
\begin{figure}[t]
  \centering
    \includegraphics[width=0.8\linewidth]{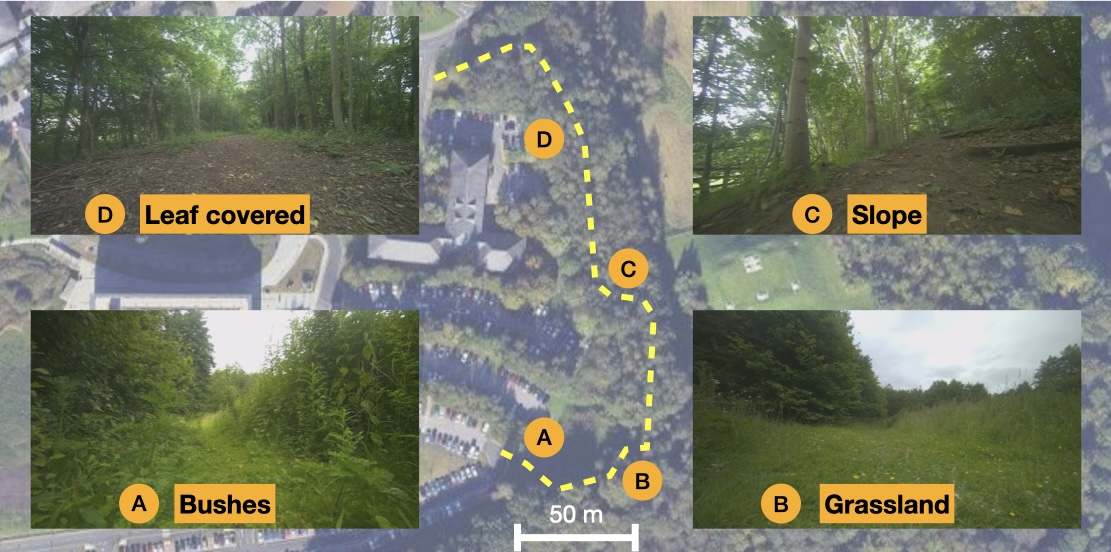}
   \caption{Our off-road trail-based data collection route (yellow curve) comprising both diverse terrains and highly variable scene illumination conditions. }
   \label{fig:routine}
\end{figure}

\noindent
Our \DODshort\ includes nine traversal sequences, collected in the hilly areas near the Department of Mathematics and Computer Science at Durham University, encompassing various natural terrains such as grasslands, bushes, trees, leaf-covered regions and slopes (see \cref{fig:routine}). We collect repeated traversal routines under varying ambient light conditions at different times of the day (July-August). The dataset comprises five sequences moving from the start point to the endpoint and four sequences in the reverse direction. The average speed was maintained at approximately 0.2 m/s, with each sequence lasting about five minutes. Each sequence is stored in ROS bag format, enabling efficient storage, synchronisation and offline analysis of recorded sensor data.

\subsection{Calibration and Synchronisation}
\label{sec:cali and sync}
\begin{figure}[t]
  \centering
    \includegraphics[width=0.8\linewidth]{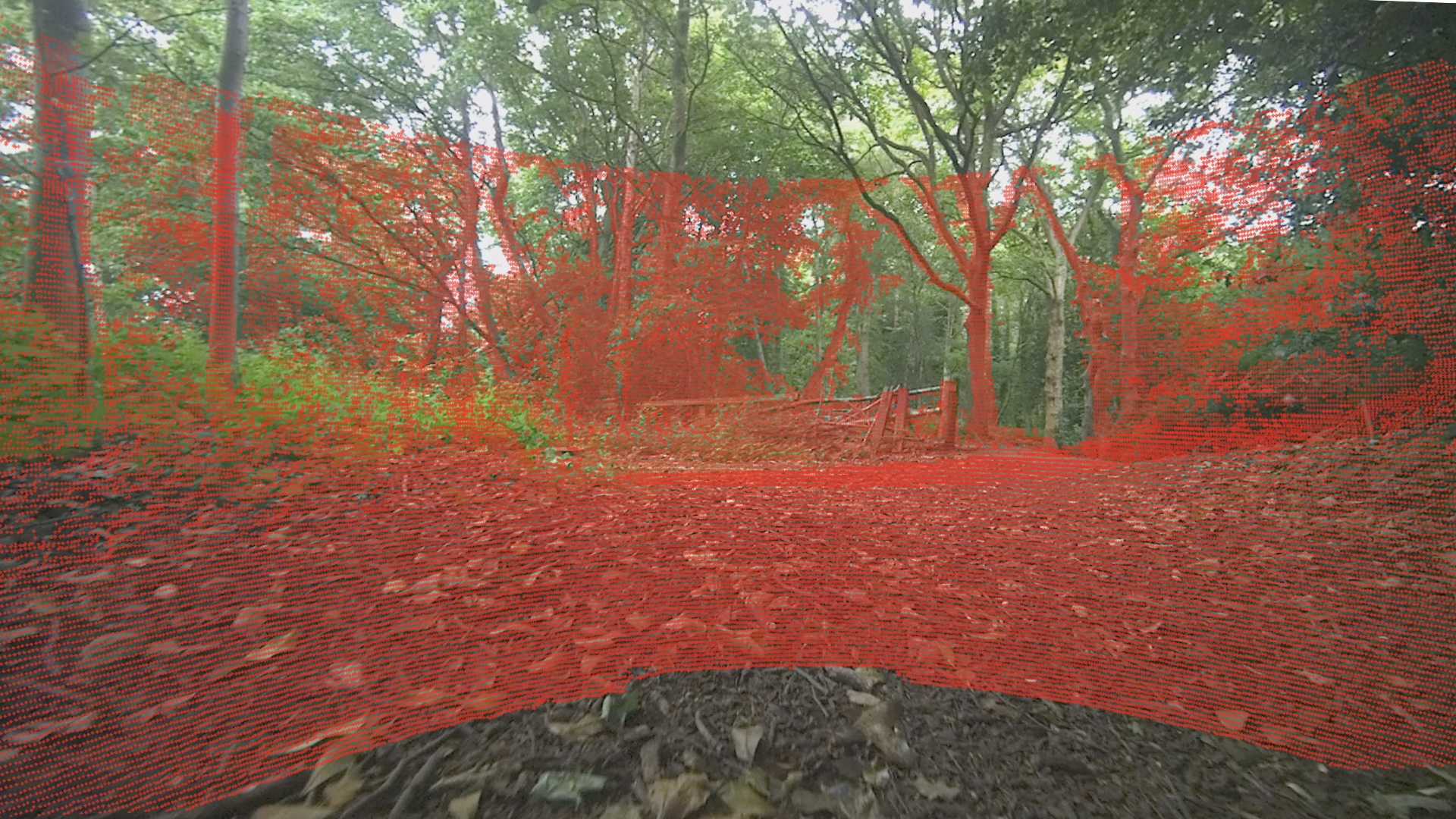}
   \caption{Evaluation of LiDAR-to-camera calibration: the 3D point cloud captured by the OS-128 LiDAR is projected onto the 2D image plane (shown in red) of the left image from the ZEDx stereo camera.}
   \label{fig:cali}
\end{figure}

\noindent
LiDAR-to-camera calibration employs a two-stage strategy. In the first stage, a target-based method \cite{guindel2017automatic} is utilised to determine the transformation matrix \([R|t]\), where \(R\) and \(t\) represent the rotation and translation parameters, respectively. In the second stage, the transformation matrix obtained from the first stage serves as a reliable initialization for a target-less method \cite{yan2022opencalib}, which further refines the calibration performance. Other sensors are strictly calibrated according to the manufacturer factory settings. \cref{fig:cali} illustrates a LiDAR to camera projection result (overlain) using the LiDAR-to-camera calibration obtained from our two-stage calibration approach. 

We implement a software-based synchronisation strategy. All sensor data is down-sampled to 10 Hz using the LiDAR frame frequency as the master. This is achieved using timestamps provided by the Robot Operating System (ROS1, Noetic).

\begin{figure*}[t] 
  \centering
    \includegraphics[width=0.9\textwidth]{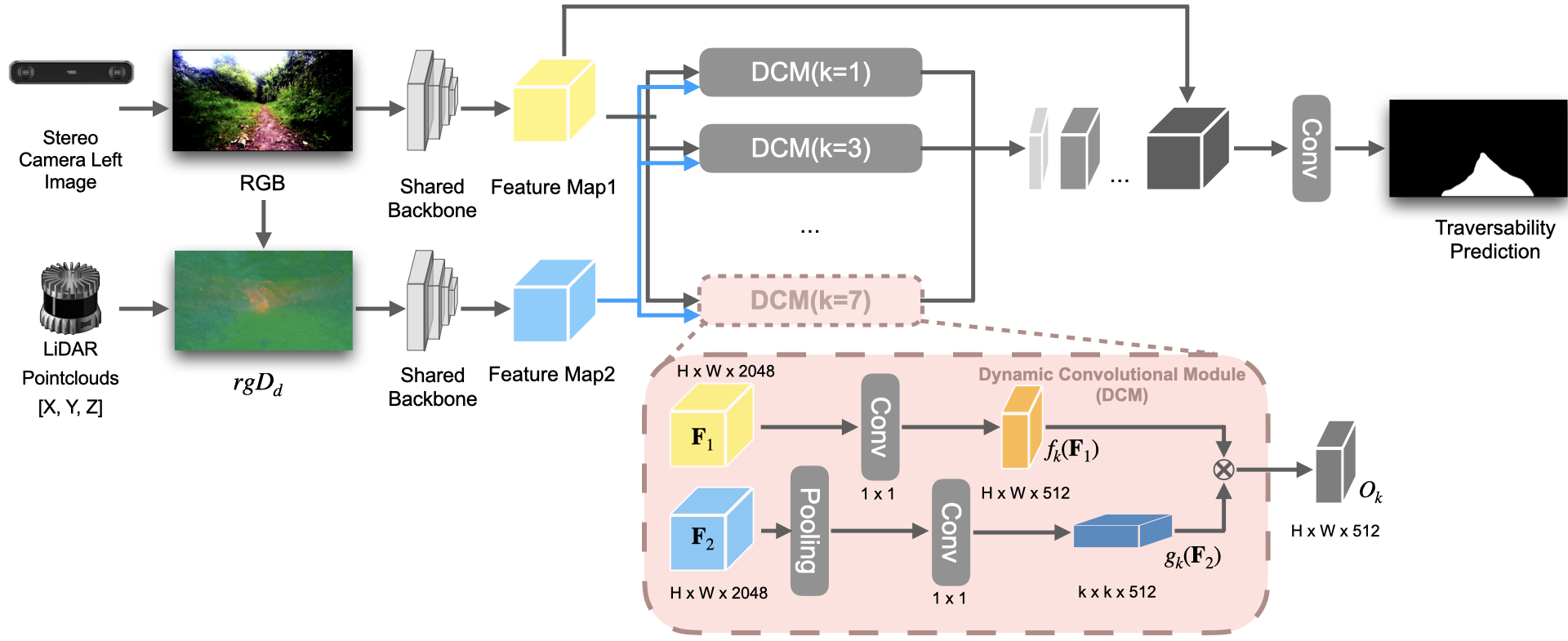} 
    \caption{Architecture of the proposed dynamic multiscale fusion network. Input modalities (e.g., RGB and $rgD_d$) are processed via a shared backbone and parallel DCMs to capture multiscale features, with context-aware filters from $F_2$ applied to $F_1$ for final prediction.}

    \label{fig:model}
\end{figure*}

\subsection{Annotation}
\noindent
Synchronized image frames are annotated to support the supervised traversable pathway segmentation task. Fast SLIC superpixels \cite{fastslic_github,achanta2012slic} are first utilized as guidance, and then subject to human annotator refinement, in order to annotate the RGB images captured from the left camera (of the ZEDx stereo camera) and label them as traversable (i.e. traversable vs. non-traversable) as a two-state binary label.

\section{Traversable Pathway Detection}
\noindent
In this section, we propose a dynamic multiscale model for cross-fusion, along with an early fusion strategy that leverages colour chromaticity to integrate visual and spatial information (i.e. camera and LiDAR data). These methods are further combined into a mixed fusion strategy. The effectiveness of each proposed fusion process is assessed under low, medium, and high ambient illumination conditions, and the evaluation results are explained in detail in Section~\ref{eva}.
\vspace{-0.2em}
\subsection{Dynamic Multiscale Data Fusion Model}
\vspace{-0.2em}
\noindent
Inspired by \cite{dynamic}, we propose a dynamic multiscale network for multi-sensor data fusion, with the architecture shown in \cref{fig:model}. The core component of the network is the Dynamic Convolutional Module (DCM), which is designed to extract multiscale feature representations in a parallel manner.
Given two feature maps generated by the backbone, 
\( \mathbf{F}_1 \in \mathbb{R}^{h \times w \times c} \) and 
\(\mathbf{F}_2 \in \mathbb{R}^{h \times w \times c}\), where \(h\), \(w\), and \(c\) represent the height, width, and number of channels of the feature maps, respectively. 

Each DCM consists of two branches. In the first branch, feature reduction \(f_k\) is applied to the input feature map \(\mathbf{F}_1\), producing a reduced feature map \(f_k(\mathbf{F}_1) \in \mathbb{R}^{h \times w \times c'}\). Here, \(c'\) is the number of channels in the reduced feature map (\(c' < c\)) and \(f_k\) is a convolution operation \(1 \times 1\) where the parameter \(k\) indicates the kernel size of the context-aware filters. Simultaneously, the second branch generates context-aware filters \( g_k(\mathbf{F}_2) \in \mathbb{R}^{k \times k \times c'} \) by applying an adaptive average pooling operation followed by a \( 1 \times 1 \) convolution operation. Subsequently, \( f_k(\mathbf{F}_1) \) is convolved with \( g_k(\mathbf{F}_2) \) using depthwise convolution, followed by a \( 1 \times 1 \) convolution, to produce the scale-specific output of the DCM (see  Eqn. \ref{eq:scale_specific_output}), where \( O_k \in \mathbb{R}^{h \times w \times c'} \):
\begin{equation}
O_k = \text{Conv}_{1 \times 1}\left(f_k(\mathbf{F}_1) \otimes g_k(\mathbf{F}_2)\right)
\label{eq:scale_specific_output}
\end{equation}

\subsection{Experimental Dataset Generation}
\label{data_gen}
\noindent
 The dataset includes an entirely annotated sequence and key frames (one in every ten frames) from three additional sequences, resulting in a total of 3,508 frames. These frames are randomly divided into training, validation and testing subsets with a split ratio of 8:1:1.

\begin{figure}[!ht]
    \centering
    \begin{subfigure}{0.15\textwidth}
        \includegraphics[width=\linewidth]{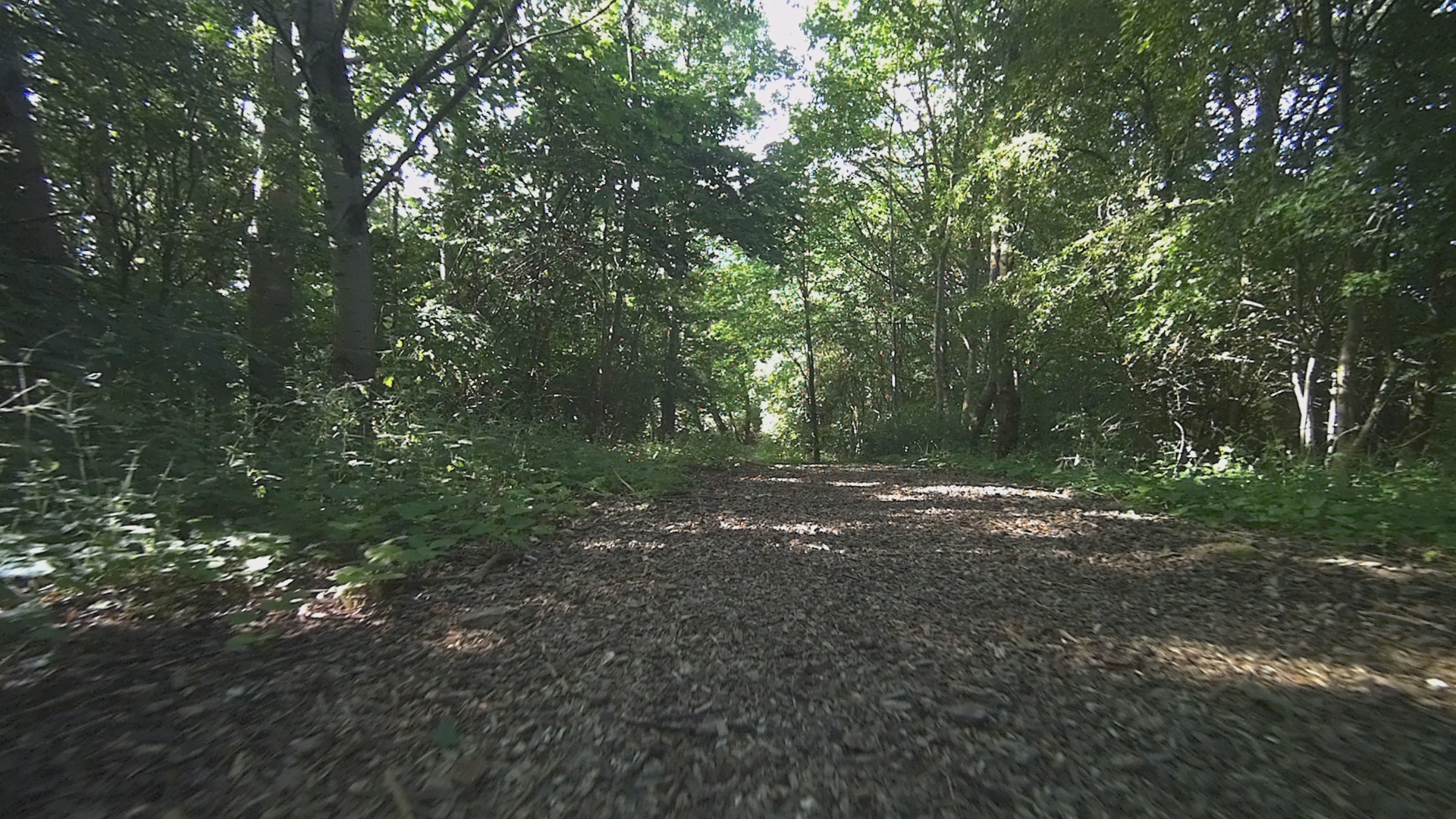}
        \caption{RGB image}
        \label{fig:sub1}
    \end{subfigure}
    \hfill
    \begin{subfigure}{0.15\textwidth}
        \includegraphics[width=\linewidth]{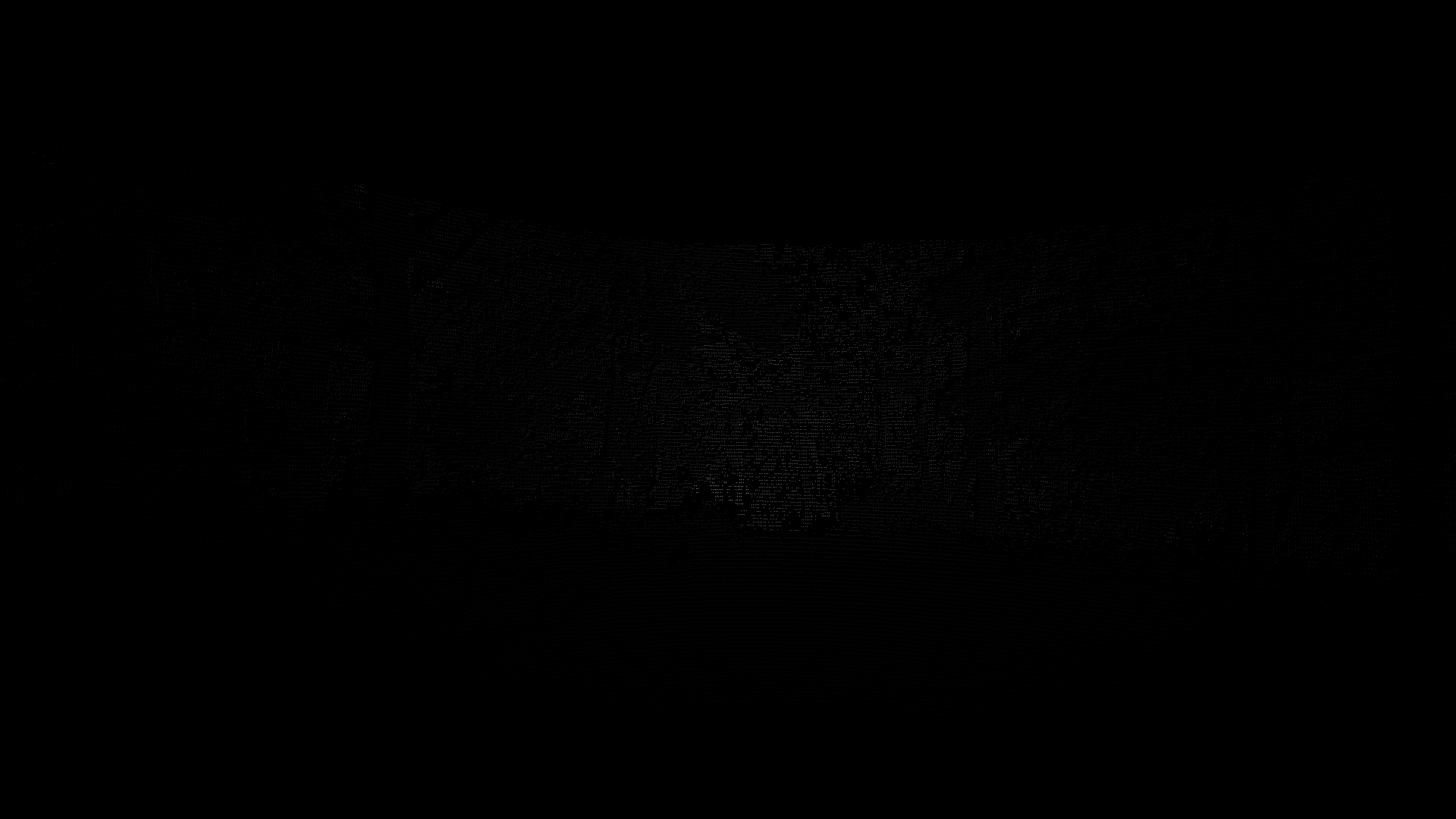}
        \caption{$D_s$}
        \label{fig:sub2}
    \end{subfigure}
    \hfill
    \begin{subfigure}{0.15\textwidth}
        \includegraphics[width=\linewidth]{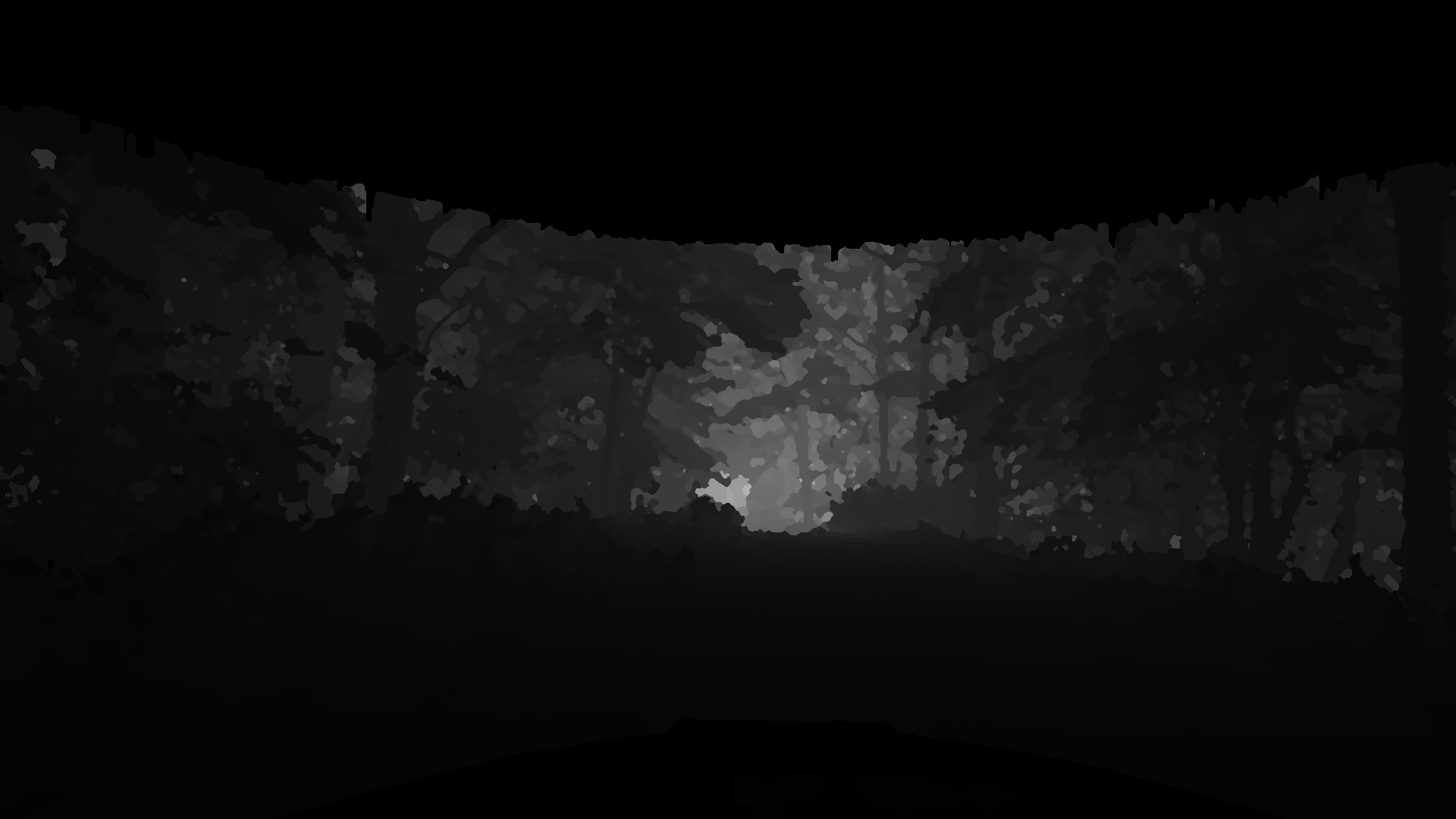}
        \caption{$D_d$}
        \label{fig:sub3}
    \end{subfigure}

    \vspace{1em}
    \begin{subfigure}{0.15\textwidth}
        \includegraphics[width=\linewidth]{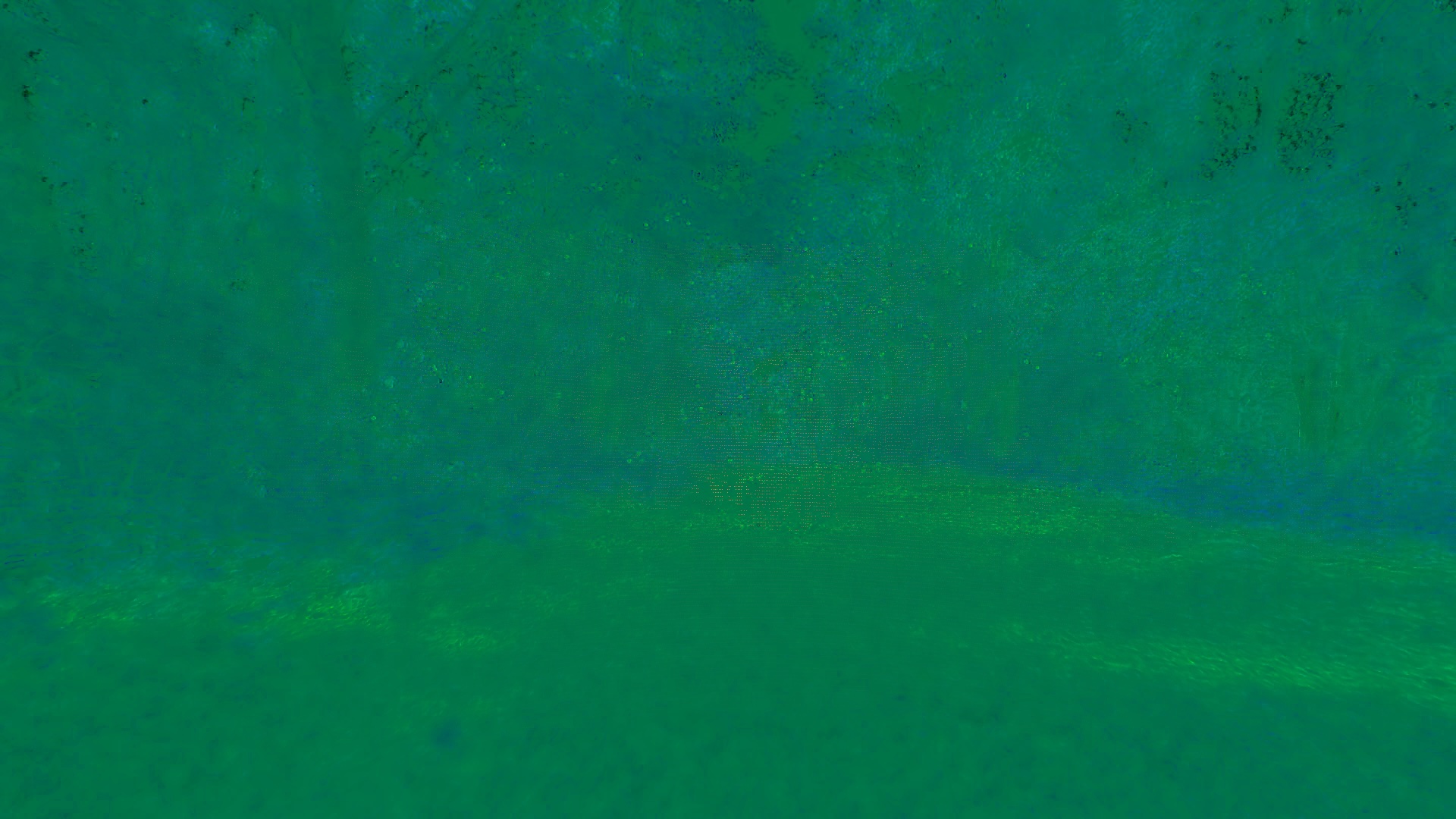}
        \caption{$rgD_s$}
        \label{fig:sub5}
    \end{subfigure}
    \hfill
    \begin{subfigure}{0.15\textwidth}
        \includegraphics[width=\linewidth]{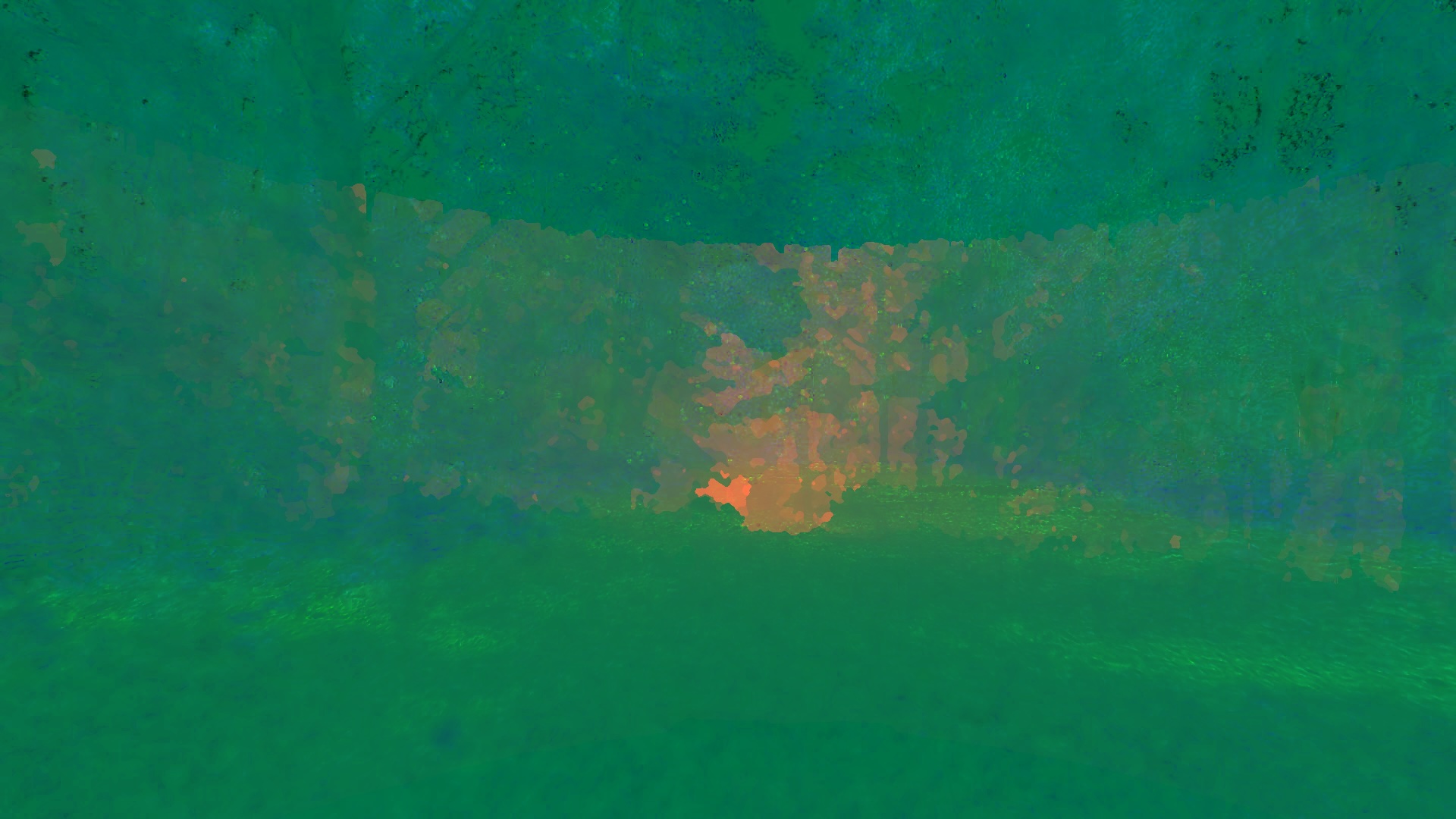}
        \caption{$rgD_d$}
        \label{fig:sub6}
    \end{subfigure}
    \hfill
    \begin{subfigure}{0.15\textwidth}
        \includegraphics[width=\linewidth]{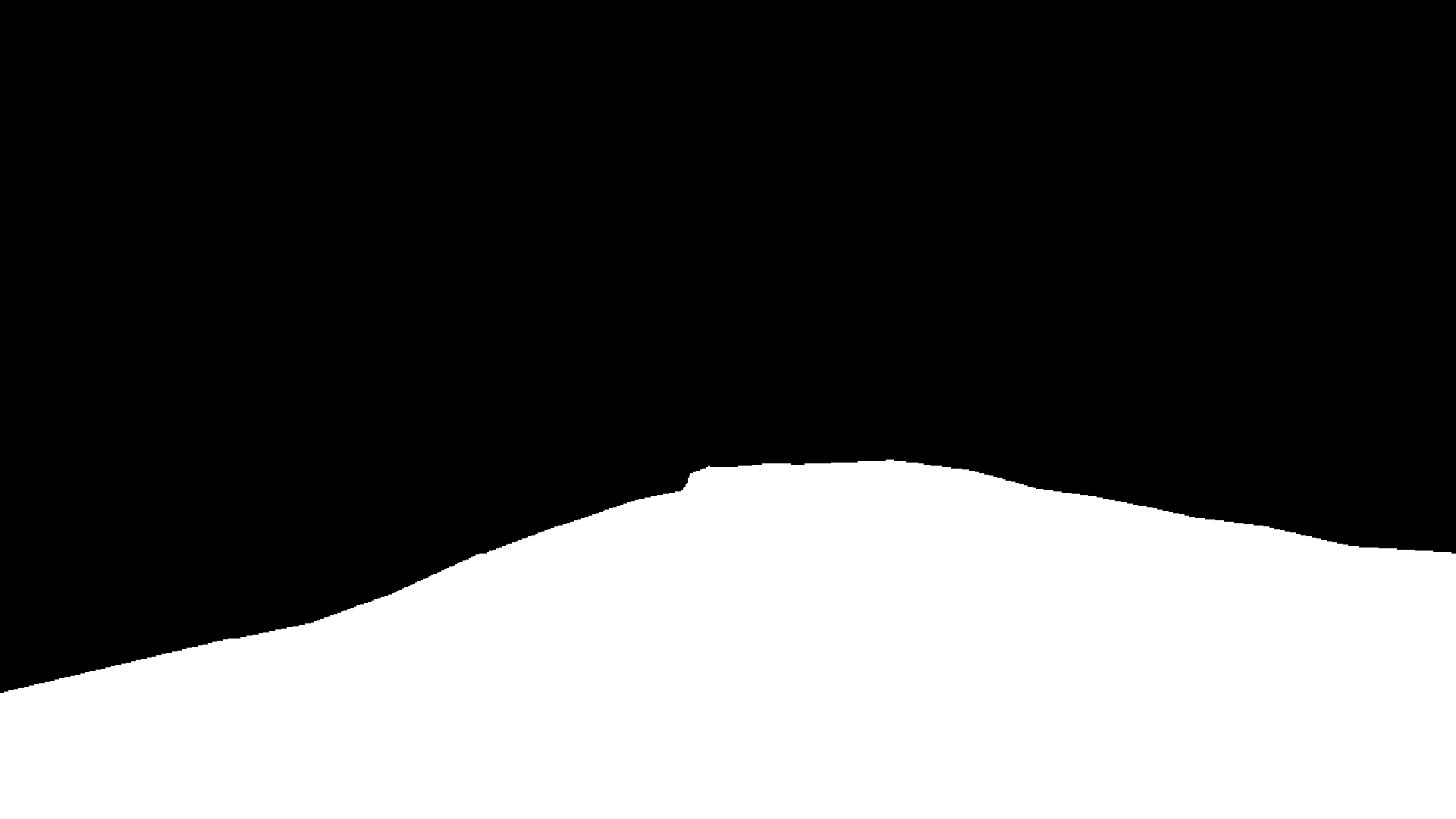}
        \caption{Annotation}
        \label{fig:sub4}
    \end{subfigure}
    \caption{An example of the five input modality combinations used in addition to the ground truth annotation of traversable pathway area.}
    \label{fig:inputs}
\end{figure}

In addition to using 2D images as the primary input modality, we incorporate corresponding sparse depth maps ($D_s$). 3D LiDAR point cloud \((X, Y, Z)\)  is firstly converted into 3D camera point cloud \((X', Y', Z')\) using:

\begin{equation}
\begin{bmatrix} X' \\ Y' \\ Z' \end{bmatrix} = R \begin{bmatrix} X \\ Y \\ Z \end{bmatrix} + t
\end{equation}

\noindent with R and t representing the extrinsic calibration matrix, obtained through the
calibration process described in Section \ref{sec:cali and sync}, between the LiDAR and the camera. 3D camera point cloud \((X', Y', Z')\) is then projected into a 2D image plane \((u, v)\) using:
\begin{equation}
\begin{aligned}
D_s(u, v) & =
\begin{cases}
Z', & \text{if } 
\begin{bmatrix} u \\ v \\ 1 \end{bmatrix} =
\operatorname{round} \left( K \begin{bmatrix} X' \\ Y' \\ Z' \end{bmatrix} \right)
- \begin{bmatrix} 1 \\ 1 \\ 0 \end{bmatrix}, \\
0, & \text{otherwise}.
\end{cases} \\
& \quad 0 \leq u < W, \quad 0 \leq v < H.
\end{aligned}
\end{equation}
\noindent with \(K\) as the intrinsic camera matrix, and $W$ and $H$ denote the width and the height of the image, respectively. For duplicate (u, v), always choose the closet point to (u, v).

We also generate dense depth maps (\(D_d\)) using a CPU-efficient depth completion method based on multi-scale dilations\cite{ku2018defense}, which incorporates additional noise removal to preserve local structure.

An early fusion approach is presented utilising colour chromaticity \cite{rg} ($r, g$) in combination with sparse or dense depth maps to ensure no additional computational burden, defined as:
\begin{equation}
\begin{bmatrix} r \\[0.1em] g \\[0.2em] b \end{bmatrix} =
\frac{1}{R + G + B} 
\begin{bmatrix} R \\[0.2em] G \\[0.2em] B \end{bmatrix}, \quad \text{for } R + G + B \neq 0.
\label{eq:chromaticity_coordinates}
\end{equation}

\mycomment{
\begin{equation}
r = \frac{R}{R + G + B}
\label{eq:chromaticity_coordinates1}
\end{equation}
\begin{equation}
g = \frac{G}{R + G + B}
\label{eq:chromaticity_coordinates2}
\end{equation}
\begin{equation}
b = \frac{B}{R + G + B}
\label{eq:chromaticity_coordinates3}
\end{equation}
}

\noindent where, $R$, $G$, and $B$ represent the red, green and blue colour channels of an original RGB image, respectively. When unit normalised, as the sum of the $(r,g,b)$ channels is unit length, discarding the $b$ channel does not result in any loss of colour information and hence removes data redundancy. The input modalities can thus be expressed as $rgD_s$ and $rgD_d$  to denote the combination of colour chromaticity, $(r,g)$ and sparse/dense depth, $D_{\{s,d\}}$. \cref{fig:inputs} illustrates an example of each type of input sensor data combination.

In order to observe the performance of the model under different ambient lighting conditions, the collected lux data is used as a metric to partition the test dataset into three ambient illuminance level subsets: low (0--100~lux),  medium (100--10,000~lux), and high (\textgreater10,000~lux). 

\subsection{Experimental Setting}
\noindent
We first compare our dynamic multi-scale data fusion model with OFF-Net \cite{min2022orfd}, and then explore the impact of different fusion strategies on segmentation performance. Following previous works\cite{holder2018encoding, min2022orfd}, We use pixel accuracy, the Intersection over Union (IoU) metric in conjunction with the F1 score to evaluate segmentation performance.

We employ Dilated Residual Networks (DRN-A-50) \cite{yu2017dilated} as the backbone of our model. The model is optimised using the Stochastic Gradient Descent with Momentum (SGDM) \cite{polyak1964some} optimiser. The initial learning rate is set to 0.001, and the batch size is configured to 40. All experiments are conducted using a Nvidia A100 GPU.

\subsection{Evaluation Results}
\label{eva}
\noindent
Table~\ref{tab:quan} presents the quantitative results on \DODshort. OFF-Net\cite{min2022orfd} uses sparse depth with linear interpolation to construct a dense depth map and employs RGB as input. It estimates surface normals to fuse with the RGB input. However, in trail-based environments, the estimated normals fluctuate drastically, leading to noisy and less reliable guidance for segmentation. Our model, which follows a mixed fusion strategy, outperforms OFF-Net\cite{min2022orfd} by +4.46\% in IoU and +2.57\% in F1-score. Additionally, our method runs significantly faster, achieving 25.58 FPS compared to OFF-Net’s 15.65 FPS, demonstrating better suitability for real-time applications.
\begin{table}[!ht]
\centering
\scriptsize
\caption{Test quantitative results on \DODshort~using metrics: accuracy, IoU, F1 score, and Frames Per Second (FPS).}
\renewcommand{\arraystretch}{1.1}
\setlength{\tabcolsep}{2pt}
\resizebox{1.0\columnwidth}{!}{%
\begin{tabular}{lllcccc}
\toprule
Model & \textbf{Input 1} & \textbf{Input 2} & \textbf{Accuracy (\%)} & \textbf{IoU (\%)} & \textbf{F1 score (\%)} & \textbf{FPS} \\
\midrule
OFF-Net\cite{min2022orfd} & $RGB$ & $D_d$ & 95.70 & 84.70 & 91.70 & 15.65 \\
\rowcolor{gray!10}
Ours & $RGB$ & $rgD_d$ & \textbf{95.85} & \textbf{89.16} & \textbf{94.27} & 25.58 \\
\bottomrule
\end{tabular}
}
\label{tab:quan}
\end{table}

\begin{figure*}[t] 
  \centering
    \includegraphics[width=1.0\textwidth]{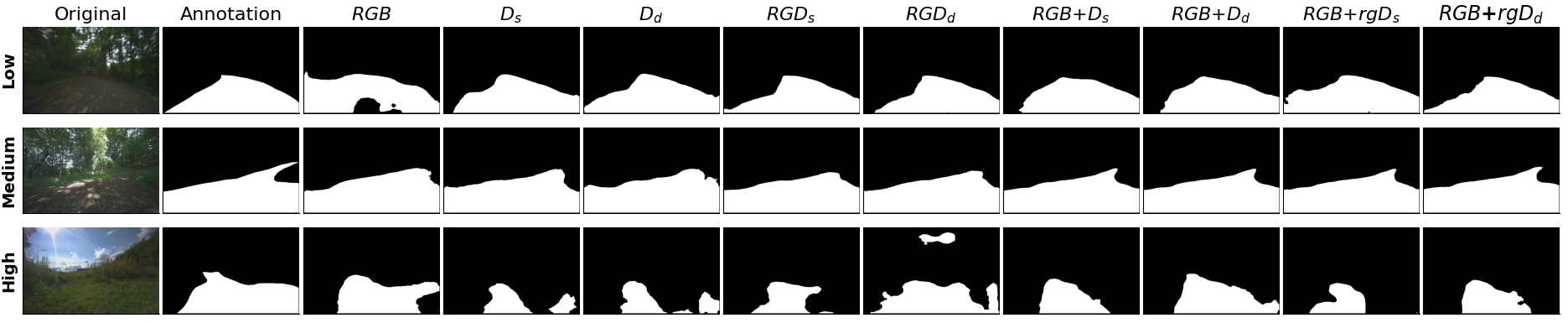} 
    \caption{Trail prediction examples corresponding to Table~\ref{tab:comparison}. The row labels represent the ambient light levels, while the column labels denote the input modalities, as explained in Section~\ref{data_gen}. The ambient illumination of the original images in the first, second, and last rows are 22.80 lx, 1512.30 lx, and 45445.10 lx, respectively.}
    \label{fig:pre}
\end{figure*}

A comprehensive comparison of model evaluation results across three illumination levels is presented in Table~\ref{tab:comparison}, alongside trail prediction examples in Fig.~\ref{fig:pre}. Under low illumination conditions (Table~\ref{tab:comparison} - (a)), the single RGB modality performs poorly due to the limited ability of the camera to capture fine details in underexposed off-road, trail-based scenarios. In contrast, depth information effectively compensates for this limitation by providing stable spatial details unaffected by lighting variations. Notably, early, cross and mixed fusion strategies, particularly when fusing dense depth, yield substantial improvements in traversable pathway segmentation performance.

At medium illumination levels (Table~\ref{tab:comparison} - (b)), cross and mixed fusion approaches offer only marginal enhancements in trail detection compared to the RGB-only input. Under high illumination conditions (Table~\ref{tab:comparison} - (c)), the most challenging scenario, frames primarily depict open grassland and similar scenes, where increased noise in depth maps may impact the cross-fusion results. This suggests that the limited adaptability of depth-generated dynamic filters to color-specific features leads to suboptimal feature integration. However, early and mixed fusion strategies with $rgD_d$ integration result in a noticeable increase in IoU, demonstrating the advantages of leveraging depth information even in well illuminated environments.

The overall test evaluation results are presented in Table~\ref{tab:comparison} - (d). Firstly, in the absence of data fusion, both sparse and dense depth inputs outperform single RGB images. Secondly, the comparable performance of sparse and dense depth maps indicates that high-resolution LiDAR data provides rich spatial information, effectively enhancing the segmentation model. Lastly, mixed fusion with RGB and $rgD_d$ achieves the best trail detection performance, with an accuracy improvement of +1.76\%, an IoU increase of +3.69\%, and an F1-score gain of +4.97\%.
\begin{table}[h]
    \centering
    \caption{Comparison of Performance Under Various Ambient Illumination Levels. The highest performance metric result in each column is highlighted in \textbf{bold}, while the second-highest performance metric result is \underline{underlined}.}
    \label{tab:comparison}
    \begin{subtable}{0.5\textwidth}
        \centering
        \small
        \caption{low (0--100~lux)}
        \resizebox{\columnwidth}{!}{
        \begin{tabular}{lllccc}
            \toprule
            \textbf{Data fusion} & \textbf{Input 1} & \textbf{Input 2} & \textbf{Accuracy (\%)} & \textbf{IoU (\%)} & \textbf{F1 score (\%)}\\
            \midrule
            N/A & $RGB$ & $RGB$ &91.06 & 80.13 &  88.46  \\
            \rowcolor{gray!10}
            Early & $rgD_s$ & $rgD_s$ & 94.51 & 86.65 & 92.36 \\
            \rowcolor{gray!10}
            Early & $rgD_d$ & $rgD_d$ & 94.65  & 87.02 & 92.72 \\
            Cross & $RGB$ & $D_s$  & 95.49 & 89.27 & 94.15 \\
            Cross & $RGB$ & $D_d$  & \textbf{95.78} & \textbf{89.99} & \textbf{94.57} \\
            \rowcolor{gray!10}
            Mixed & $RGB$ & $rgD_s$  & 95.32 & 89.02 & 94.01 \\
            \rowcolor{gray!10}
            Mixed & $RGB$ & $rgD_d$  & \underline{95.62} & \underline{89.62} & \underline{94.37} \\
            \bottomrule
        \end{tabular}
        }
        \vspace{0.2cm}
    \end{subtable}
    \begin{subtable}{0.5\textwidth}
        \centering
        \small
        \caption{medium (100--10,000~lux)}
        \resizebox{\columnwidth}{!}{
        \begin{tabular}{lllccc}
            \toprule
            \textbf{Data fusion} & \textbf{Input 1} & \textbf{Input 2} & \textbf{Accuracy (\%)} & \textbf{IoU (\%)} & \textbf{F1 score (\%)}\\
            \midrule
            
            N/A & $RGB$ & $RGB$ & 96.71 & 91.71 & 95.36 \\
            \rowcolor{gray!10}
            Early & $rgD_s$ & $rgD_s$ & 96.54 & 90.95 & 94.67 \\
            \rowcolor{gray!10}
            Early & $rgD_d$ & $rgD_d$ & 95.77 & 89.88 & 94.40 \\
            Cross & $RGB$ & $D_s$ & \textbf{97.61} & \underline{93.71} & \underline{96.56}\\
            Cross & $RGB$ & $D_d$ &  \underline{97.60} & \textbf{93.79} & \textbf{96.63} \\
            \rowcolor{gray!10}
            Mixed & $RGB$ & $rgD_s$ & 97.24 & 92.87 & 96.11 \\
            \rowcolor{gray!10}
            Mixed & $RGB$ & $rgD_d$ & 97.19 & 93.05 & 96.17 \\
            \bottomrule
        \end{tabular}        }
        \vspace{0.2cm}
    \end{subtable}
    
    \begin{subtable}{0.5\textwidth}
        \centering
        \small
        \caption{high (\textgreater10,000~lux)}
        \resizebox{\columnwidth}{!}{
        \begin{tabular}{lllccc}
            \toprule
            \textbf{Data fusion} & \textbf{Input 1} & \textbf{Input 2} & \textbf{Accuracy (\%)} & \textbf{IoU (\%)} & \textbf{F1 score (\%)}\\
            \midrule
            N/A & $RGB$ & $RGB$  & 92.68 & 79.58 & 79.58 \\
            \rowcolor{gray!10}
            Early & $rgD_s$ & $rgD_s$ & 91.09  & 75.34 & 84.33 \\
            \rowcolor{gray!10}
            Early & $rgD_d$ & $rgD_d$ & \underline{92.71}  & \underline{79.88} & \underline{87.95} \\
            
            Cross & $RGB$ & $D_s$  & 90.88 & 73.19 & 80.60 \\
            Cross & $RGB$ & $D_d$  & 91.52 & 75.40 & 83.05 \\
            \rowcolor{gray!10}
            Mixed & $RGB$ & $rgD_s$ & 91.50 & 75.48 & 83.11 \\
            \rowcolor{gray!10}
            Mixed & $RGB$ & $rgD_d$ & \textbf{93.76} & \textbf{82.01} & \textbf{88.90} \\  
            \bottomrule
        \end{tabular}
        }
        \vspace{0.2cm}
    \end{subtable}
    \begin{subtable}{0.5\textwidth}
        \centering
        \small
        \caption{Summary}
        \resizebox{\columnwidth}{!}{
        \begin{tabular}{lllccc}
            \toprule
            \textbf{Data fusion} & \textbf{Input 1} & \textbf{Input 2} & \textbf{Accuracy (\%)} & \textbf{IoU (\%)} & \textbf{F1 score (\%)}\\
            \midrule
            N/A & $RGB$ & $RGB$ & 94.18 & 85.47 & 89.30  \\
    
            N/A & $D_s$ & $D_s$ & 94.74 & 86.10 & 91.72 \\
            N/A & $D_d$ & $D_d$ & 94.91 & 86.52 & 92.09 \\
            \rowcolor{gray!10}
            Early & $rgD_s$ & $rgD_s$ & 94.53 & 85.59 & 91.25 \\
            \rowcolor{gray!10}
            Early & $rgD_d$ & $rgD_d$ & 95.07  & 87.18 & \underline{92.59} \\

            Cross & $RGB$ & $D_s$  & 95.30 & 87.18 & 91.72 \\
            Cross & $RGB$ & $D_d$  & \underline{95.41} & \underline{87.57} & 92.25 \\
        
            \rowcolor{gray!10}
            Mixed & $RGB$ & $rgD_s$  & 95.24 & 87.23 & 92.09 \\
            \rowcolor{gray!10}
            Mixed & $RGB$ & $rgD_d$  & \textbf{95.85} & \textbf{89.16} & \textbf{94.27} \\
            \bottomrule
        \end{tabular}
        }
    \end{subtable}
\end{table}
\noindent

\section{conclusion}
\noindent

We propose the \DODshort\ dataset, specifically designed for unstructured and complex trail-like scenarios using a medium-scale all-terrain robot platform. The dataset includes high-fidelity multimodal sensor data, such as 128-channel 3D LiDAR, stereo imagery, GNSS, IMU, telemetry control data, and illumination measurements, collected through repeated route traversals under varying environmental conditions. It comprises 31.4k frame pairs of image and LiDAR, with annotated traversability levels for key frames. Furthermore, we propose a novel dynamic multi-scale data fusion model for precise traversable trail-like area prediction. The evaluation of early, cross, and mixed fusion processes under different illumination conditions highlights their influence on model performance and demonstrates the potential applicability the dataset across diverse environmental settings.

Future work will expand the dataset with diverse routes featuring varying vegetation densities and natural obstacles, such as rocks and water crossings. We plan to capture seasonal and weather variations and provide fine-grained annotations such as obstacle types and surface roughness for terrain classification and adaptive navigation. Furthermore, we will integrate temporal information for dynamic changes and explore attention-based mechanisms for better feature fusion. Finally, self-supervised learning will be employed to reduce annotation efforts and improve performance in underrepresented scenarios.


\vspace{-0.5em}
{\scriptsize
\bibliographystyle{IEEEtran}
\bibliography{egbib}

\begin{thebibliography}{10}
\providecommand{\url}[1]{#1}
\csname url@samestyle\endcsname
\providecommand{\newblock}{\relax}
\providecommand{\bibinfo}[2]{#2}
\providecommand{\BIBentrySTDinterwordspacing}{\spaceskip=0pt\relax}
\providecommand{\BIBentryALTinterwordstretchfactor}{4}
\providecommand{\BIBentryALTinterwordspacing}{\spaceskip=\fontdimen2\font plus
\BIBentryALTinterwordstretchfactor\fontdimen3\font minus \fontdimen4\font\relax}
\providecommand{\BIBforeignlanguage}[2]{{%
\expandafter\ifx\csname l@#1\endcsname\relax
\typeout{** WARNING: IEEEtran.bst: No hyphenation pattern has been}%
\typeout{** loaded for the language `#1'. Using the pattern for}%
\typeout{** the default language instead.}%
\else
\language=\csname l@#1\endcsname
\fi
#2}}
\providecommand{\BIBdecl}{\relax}
\BIBdecl

\bibitem{geiger2013vision}
A.~Geiger, P.~Lenz, C.~Stiller, and R.~Urtasun, ``Vision meets robotics: The kitti dataset,'' \emph{Int. J. Comput. Vis. (IJCV)}, vol.~32, no.~11, pp. 1231--1237, 2013.

\bibitem{sun2020scalability}
P.~Sun, H.~Kretzschmar, X.~Dotiwalla, A.~Chouard, V.~Patnaik \emph{et~al.}, ``Scalability in perception for autonomous driving: Waymo open dataset,'' in \emph{Proc. IEEE/CVF Conf. Comput. Vis. Pattern Recognit. (CVPR)}, 2020, pp. 2446--2454.

\bibitem{caesar2020nuscenes}
H.~Caesar, V.~Bankiti, A.~H. Lang \emph{et~al.}, ``nuscenes: A multimodal dataset for autonomous driving,'' in \emph{Proc. IEEE/CVF Conf. Comput. Vis. Pattern Recognit. (CVPR)}, 2020, pp. 11\,621--11\,631.

\bibitem{SOOR}
O.~Mayuku, B.~W. Surgenor, and J.~A. Marshall, ``Multi-resolution and multi-domain analysis of off-road datasets for autonomous driving,'' in \emph{Proc. Conf. Robot. Vis. (CRV)}, 2021, pp. 165--172.

\bibitem{wildscenes}
K.~Vidanapathirana, J.~Knights, S.~Hausler, M.~Cox \emph{et~al.}, ``Wildscenes: A benchmark for 2d and 3d semantic segmentation in large-scale natural environments,'' \emph{Int. J. Robot. Res.}, 2024.

\bibitem{YCOR}
D.~Maturana, P.-W. Chou, M.~Uenoyama, and S.~Scherer, ``Real-time semantic mapping for autonomous off-road navigation,'' in \emph{Field and Service Robotics (FSR)}.\hskip 1em plus 0.5em minus 0.4em\relax Springer, 2018, pp. 335--350.

\bibitem{min2022orfd}
C.~Min, W.~Jiang, D.~Zhao, J.~Xu, L.~Xiao, Y.~Nie, and B.~Dai, ``Orfd: A dataset and benchmark for off-road freespace detection,'' in \emph{Proc. IEEE Int. Conf. Robot. Autom. (ICRA)}.\hskip 1em plus 0.5em minus 0.4em\relax IEEE, 2022, pp. 2532--2538.

\bibitem{CaT}
S.~Sharma, L.~Dabbiru, T.~Hannis, G.~Mason, D.~W. Carruth, M.~Doude, C.~Goodin, C.~Hudson, S.~Ozier, J.~E. Ball, and B.~Tang, ``Cat: Cavs traversability dataset for off-road autonomous driving,'' \emph{IEEE Access}, vol.~10, pp. 24\,759--24\,768, 2022.

\bibitem{goose}
P.~Mortimer, R.~Hagmanns, M.~Granero, T.~Luettel, J.~Petereit, and H.-J. Wuensche, ``The goose dataset for perception in unstructured environments,'' in \emph{Proc. IEEE Int. Conf. Robot. Autom. (ICRA)}, 2024, pp. 14\,838--14\,844.

\bibitem{tartandrive}
M.~Sivaprakasam, P.~Maheshwari, M.~G. Castro, S.~Triest, M.~Nye, S.~Willits, A.~Saba, W.~Wang, and S.~Scherer, ``Tartandrive 2.0: More modalities and better infrastructure to further self-supervised learning research in off-road driving tasks,'' in \emph{Proc. IEEE Int. Conf. Robot. Autom. (ICRA)}.\hskip 1em plus 0.5em minus 0.4em\relax IEEE, 2024.

\bibitem{FreiburgForest}
A.~Valada, G.~L. Oliveira, T.~Brox, and W.~Burgard, ``Deep multispectral semantic scene understanding of forested environments using multimodal fusion,'' in \emph{Proc. Int. Symp. Exp. Robot. (ISER)}.\hskip 1em plus 0.5em minus 0.4em\relax Springer, 2017, pp. 465--477.

\bibitem{RUGD}
M.~Wigness, S.~Eum, J.~G. Rogers, D.~Han, and H.~Kwon, ``A rugd dataset for autonomous navigation and visual perception in unstructured outdoor environments,'' in \emph{Proc. IEEE/RSJ Int. Conf. Intell. Robots Syst. (IROS)}, 2019, pp. 5000--5007.

\bibitem{RELLIS-3D}
P.~Jiang, P.~Osteen, M.~Wigness, and S.~Saripalli, ``Rellis-3d dataset: Data, benchmarks and analysis,'' in \emph{Proc. IEEE Int. Conf. Robot. Autom. (ICRA)}, 2021, pp. 1110--1116.

\bibitem{botanicgarden}
Y.~Liu, Y.~Fu, M.~Qin, Y.~Xu \emph{et~al.}, ``Botanicgarden: A high-quality dataset for robot navigation in unstructured natural environments,'' \emph{IEEE Robot. Autom. Lett.}, 2024.

\bibitem{finnwoodlands}
J.~Lagos, U.~Lempi{\"o}, and E.~Rahtu, ``Finnwoodlands dataset,'' in \emph{Scand. Conf. Image Anal. (SCIA)}.\hskip 1em plus 0.5em minus 0.4em\relax Springer, 2023, pp. 95--110.

\bibitem{li2021durlar}
L.~Li, K.~N. Ismail, H.~P.~H. Shum, and T.~P. Breckon, ``Durlar: A high-fidelity 128-channel lidar dataset with panoramic ambient and reflectivity imagery,'' in \emph{Proc. Int. Conf. 3D Vis. (3DV)}, 2021, pp. 1227--1237.

\bibitem{bauer2001relevance}
A.~Bauer, K.~Dietz, G.~Kolling, W.~Hart, and U.~Schiefer, ``The relevance of stereopsis for motorists: A pilot study,'' \emph{Graefe’s Arch. Clin. Exp. Ophthalmol.}, vol. 239, pp. 400--406, 2001.

\bibitem{liang2022bevfusion}
T.~Liang, H.~Xie, K.~Yu \emph{et~al.}, ``Bevfusion: A simple and robust lidar-camera fusion framework,'' \emph{Adv. Neural Inf. Process. Syst.}, vol.~35, pp. 10\,421--10\,434, 2022.

\bibitem{xiao2018hybrid}
L.~Xiao, R.~Wang, B.~Dai \emph{et~al.}, ``Hybrid conditional random field-based camera-lidar fusion for road detection,'' \emph{Inf. Sci.}, vol. 432, pp. 543--558, 2018.

\bibitem{holder2018encoding}
C.~J. Holder and T.~P. Breckon, ``Encoding stereoscopic depth features for scene understanding in off-road environments,'' in \emph{Proc. Int. Conf. Image Anal. Recognit. (ICIAR)}, 2018, pp. 427--434.

\bibitem{chen2019progressive}
Z.~Chen, J.~Zhang, and D.~Tao, ``Progressive lidar adaptation for road detection,'' \emph{IEEE/CAA J. Autom. Sin.}, vol.~6, no.~3, pp. 693--702, 2019.

\bibitem{gu2021cascaded}
S.~Gu, J.~Yang, and H.~Kong, ``A cascaded lidar-camera fusion network for road detection,'' in \emph{Proc. IEEE Int. Conf. Robot. Autom. (ICRA)}, 2021, pp. 13\,308--13\,314.

\bibitem{caltagirone2019lidar}
L.~Caltagirone, M.~Bellone, L.~Svensson, and M.~Wahde, ``Lidar-camera fusion for road detection using fully convolutional neural networks,'' \emph{Robot. Auton. Syst.}, vol. 111, pp. 125--131, 2019.

\bibitem{holder2016road}
C.~J. Holder, T.~P. Breckon, and X.~Wei, ``From on-road to off: Transfer learning within a deep cnn for off-road scene classification,'' in \emph{Proc. Eur. Conf. Comput. Vis. Workshops (ECCV)}, 2016, pp. 149--162.

\bibitem{guindel2017automatic}
C.~Guindel, J.~Beltrán, D.~Martín, and F.~García, ``Automatic extrinsic calibration for lidar-stereo vehicle sensor setups,'' in \emph{Proc. IEEE Intell. Transp. Syst. Conf. (ITSC)}, 2017, pp. 1--6.

\bibitem{yan2022opencalib}
G.~Yan, Z.~Liu, C.~Wang \emph{et~al.}, ``Opencalib: A multi-sensor calibration toolbox for autonomous driving,'' \emph{Softw. Impacts}, vol.~14, p. 100393, 2022.

\bibitem{fastslic_github}
Algy, ``Fast slic: A fast, memory efficient implementation of slic superpixel segmentation,'' \url{https://github.com/Algy/fast-slic}, 2025, accessed: 2025-01-27.

\bibitem{achanta2012slic}
R.~Achanta, A.~Shaji, K.~Smith \emph{et~al.}, ``Slic superpixels compared to state-of-the-art superpixel methods,'' \emph{IEEE Trans. Pattern Anal. Mach. Intell.}, vol.~34, no.~11, pp. 2274--2282, 2012.

\bibitem{dynamic}
J.~He, Z.~Deng, and Y.~Qiao, ``Dynamic multi-scale filters for semantic segmentation,'' in \emph{Proc. IEEE/CVF Int. Conf. Comput. Vis. (ICCV)}, 2019, pp. 3562--3572.

\bibitem{ku2018defense}
J.~Ku, A.~Harakeh, and S.~L. Waslander, ``In defense of classical image processing: Fast depth completion on the cpu,'' in \emph{Proc. IEEE Conf. Robot. Vis. (CRV)}, 2018, pp. 16--22.

\bibitem{rg}
S.~Ratnasingam and T.~M. McGinnity, ``Chromaticity space for illuminant invariant recognition,'' \emph{IEEE Trans. Image Process. (TIP)}, vol.~21, no.~8, pp. 3612--3623, 2012.

\bibitem{yu2017dilated}
F.~Yu, V.~Koltun, and T.~Funkhouser, ``Dilated residual networks,'' in \emph{Proc. IEEE/CVF Conf. Comput. Vis. Pattern Recognit. (CVPR)}, 2017, pp. 472--480.

\bibitem{polyak1964some}
B.~T. Polyak, ``Some methods of speeding up the convergence of iteration methods,'' \emph{USSR Comput. Math. Math. Phys.}, vol.~4, no.~5, pp. 1--17, 1964.

\end{thebibliography}
}
\vspace{-2em}

\end{document}